\def\year{2022}\relax
\documentclass[letterpaper]{article} % DO NOT CHANGE THIS
\usepackage{aaai22}  % DO NOT CHANGE THIS
\usepackage{times}  % DO NOT CHANGE THIS
\usepackage{helvet}  % DO NOT CHANGE THIS
\usepackage{courier}  % DO NOT CHANGE THIS
\usepackage[hyphens]{url}  % DO NOT CHANGE THIS
\usepackage{graphicx} % DO NOT CHANGE THIS
\usepackage{multirow}

\usepackage{natbib}  % DO NOT CHANGE THIS AND DO NOT ADD ANY OPTIONS TO IT
\usepackage{caption} % DO NOT CHANGE THIS AND DO NOT ADD ANY OPTIONS TO IT
\DeclareCaptionStyle{ruled}{labelfont=normalfont,labelsep=colon,strut=off} % DO NOT CHANGE THIS
\usepackage{algorithm}
\usepackage{algorithmic}
\usepackage{amsmath}
\usepackage{threeparttable}
\usepackage{amssymb}
\usepackage{newfloat}
\usepackage{listings}
\floatstyle{ruled}
\newfloat{listing}{tb}{lst}{}
\floatname{listing}{Listing}
\title{Backpropagation with Biologically Plausible Spatio-Temporal Adjustment \\For Training Deep Spiking Neural Networks }
\author{
    Guobin Shen\textsuperscript{\rm 1,2,3}\equalcontrib,
    Dongcheng Zhao\textsuperscript{\rm 1,3} \equalcontrib,
    Yi Zeng\textsuperscript{\rm 1,2,3,4,5}\thanks{Corresponding Author.}
}
\affiliations{
    \textsuperscript{\rm 1}Institute of Automation, Chinese Academy of Sciences (CASIA), Beijing, China\\
    \textsuperscript{\rm 2}School of Future Technology, University of Chinese Academy of Sciences, Beijing, China\\
    \textsuperscript{\rm 3}Research Center for Brain-Inspired Intelligence, CASIA, Beijing, China\\
    \textsuperscript{\rm 4}National Laboratory of Pattern Recognition, CASIA, Beijing, China\\
    \textsuperscript{\rm 5}Center for Excellence in Brain Science and Intelligence Technology, \\Chinese Academy of Sciences, Shanghai, China \\

    \{shenguobin2021, zhaodongcheng2016, yi.zeng\}@ia.ac.cn
}

\usepackage{bibentry}
\usepackage{booktabs}
\begin{document}

\maketitle

\begin{abstract}
The spiking neural network (SNN) mimics the information processing operation in the human brain, represents and transmits information in spike trains containing wealthy spatial and temporal information, and shows superior performance on many cognitive tasks. In addition, the event-driven information processing enables the energy-efficient implementation on neuromorphic chips. The success of deep learning is inseparable from backpropagation. Due to the discrete information transmission, directly applying the backpropagation to the training of the SNN still has a performance gap compared with the traditional deep neural networks. Also, a large simulation time is required to achieve better performance, which results in high latency. To address the problems, we propose a biological plausible spatial adjustment, which rethinks the relationship between membrane potential and spikes and realizes a reasonable adjustment of gradients to different time steps. And it precisely controls the backpropagation of the error along the spatial dimension. Secondly, we propose a biologically plausible temporal adjustment making the error propagate across the spikes in the temporal dimension, which overcomes the problem of the temporal dependency within a single spike period of the traditional spiking neurons. We have verified our algorithm on several datasets, and the experimental results have shown that our algorithm greatly reduces the network latency and energy consumption while also improving network performance. We have achieved state-of-the-art performance on the neuromorphic datasets N-MNIST, DVS-Gesture, and DVS-CIFAR10. For the static datasets MNIST and CIFAR10, we have surpassed most of the traditional SNN backpropagation training algorithm and achieved relatively superior performance.  
\end{abstract}

\section{Introduction}
Deep neural networks have achieved success in various research areas, such as object detection~\citep{zou2019object}, visual tracking~\citep{li2018deep} and face recognition~\cite{masi2018deep}, etc. However, they are still far away from the information processing mechanisms of the human brain. Spiking neural networks (SNNs) are known as the third generation artificial neural network~\citep{maass1997networks}, which utilizes the discrete spikes to transmit information 
As a result, they are more energy-efficient and more in line with the information processing method in the biological brain.

However, due to the complex neural dynamics and non-differential characteristics of SNNs, it is still a challenge to train SNNs efficiently. Existing SNNs training methods can be roughly divided into three categories: the biologically plausible, conversion, and backpropagation-based strategies.

For the biologically plausible method, which is mainly inspired by the synaptic learning rules in the human brains, such as Hebbian learning rules~\citep{hebb1949organization} and Spike-Timing-Dependent Plasticity (STDP)~\cite{bi1998synaptic}. The Hebbian theory believes that the connection between pre-synaptic neurons to post-synaptic neurons will increase due to continuous and repetitive stimulation of pre-synaptic neurons. STDP is an extended Hebbian learning rule based on the temporal difference between pre and post-synaptic neurons. Diehl~\citep{diehl2015unsupervised} used the STDP learning rule and lateral inhibition in two-layer SNNs and achieved 95\% accuracy on the MNIST dataset. Saeed et al.~\citep{kheradpisheh2016bio} introduced a weight sharing strategy and designed a spiking convolutional neural network. They use STDP to learn the network weights layer by layer. Kherapisheh et al.~\citep{kheradpisheh2018stdp} used the hand-crafted DoG (Difference of Gaussian) features as the input of the SNNs, and trained the subsequently convolutional layer through STDP. However, these methods rely on the local activities of neighboring neurons to update network weights and lack the supervision of global signals. Although GLSNN~\citep{zhao2020glsnn} introduced the global feedback connections to transmit global information, it still performs poorly when transplanted to some deep networks for some complex tasks.

And the conversion method is an alternative way to get high-performance SNNs. The conversion method first trains a DNN, then converts the well-trained DNNs to SNNs with the same structure. The analog value of DNNs is converted into the firing rates of SNNs with some additional adjustments to SNNs~\cite{diehl2015fast,xu2018csnn,sengupta2019going, Hu2018SpikingDR}. Although the conversion method makes the SNNs achieve performance close to the traditional DNNs, the simulation time is too long, which causes the network to have poor real-time performance and high energy consumption. Also, the conversion methods rely highly on the well-trained DNNs and do not take advantage of the temporal information of SNNs. 

%\begin{figure}[htbp]
%\centering
%\includegraphics[width=8cm]{lif_prob.pdf}
%\caption{The backpropagation process of SNNs. The surrogate gradient will propagate gradient through unspiking neurons. The temporal dependency of BP is truncated by the reset mechanism in the LIF neurons.  }
%\label{fig_lif_prob}
%\end{figure} 

The success of deep learning is due to the proposal of the backpropagation algorithm. Subsequently, researchers in SNNs domains also tried to introduce the backpropagation algorithm into the optimization of SNNs using surrogate gradient method~\cite{  wu2018spatio,wu2019direct,jin2018hybrid,zhang2020temporal}. Surrogate gradient helps SNNs perform backpropagation through time (BPTT) so that SNNs can be adapted to larger-scale network structures, such as VGG16, ResNet, etc., and perform better on more complex ones datasets. In recent years, many spiking neuron models with biological neural characteristics have been proposed, and the Leaky-Integrate-and-Fire (LIF)  model is adopted in most common neuron models in deep spiking neural networks. The LIF Neurons continuously accumulate the membrane potential and emit spikes once they reach the threshold, and the surrogate gradient smoothes the spike firing function to get the derivatives. And these will lead to some problems. First, the surrogate gradient makes the spiking neurons transfer errors around the threshold when it propagates back. This will cause the neurons that do not emit spikes to participate in the gradient calculation repeatedly so that neurons that emit spikes earlier have a greater impact on the weight; At the same time, LIF neurons can only propagate errors back in a single spike cycle. When the neuron reaches the threshold, the temporal dependence will be truncated after the spikes are fired. These will be discussed in detail in the background section. To address the problems mentioned above, we improved backpropagation with biologically plausible spatio-temporal adjustment, which can be summarized as follows:

\begin{itemize}
\item We study the influence of the surrogate gradient on the spatio-temporal dimension of the SNNs, rethink the relationship between the neuron membrane potential and the spikes and propose a more biological spatial adjustment to help regulate the spike activities. 
\item We study the limitations of surrogate gradient in the temporal dimension, and introduce a more biological temporal residual mechanism, which enables the SNNs to propagate back across the spikes, enhancing the temporal dependence of the SNNs. 
\item We experiment on several famous datasets. For the static datasets MNIST and CIFAR10, we get remarkable performance compared to other state-of-the-art SNNs. To our best knowledge, we have reached state-of-the-art performance for the neuromorphic datasets, N-MNIST, DVS-CIFAR10, and DVS-Gesture. Moreover, our method dramatically reduces the energy consumption and latency through analysis compared to other state-of-the-art SNNs. 
\end{itemize}

\section{Background}
In this section, we will first introduce the spiking neurons used in this paper. Then we will review the problems of the traditional spatio-temporal backpropagation algorithm used to train SNNs. 
\subsection{Spiking Neuron Model}
We give a detailed description of the LIF neuron models. As shown in Eq~\ref{lif1}, the membrane potential of the neuron changes dynamically with the input current.
\begin{equation}
	\tau \frac{du_i^l(t)}{dt} = -u_i^l(t) + RI_i^l(t) 
	\label{lif1}
\end{equation}
$I_i^l(t)$ denotes the input current, composed of input spikes.$R$ is the membrane resistance, $\tau$ is the synaptic time constant. When the membrane potential is greater than the threshold $u_{th}$, the neuron will spike and be reset to $u_{reset}$. Without loss of generality, we set the reset potential $u_{reset} = 0$. To facilitate the calculation and simulation, we convert Eq.~\ref{lif1} into a discrete form so that we can get Eq.~\ref{lif2}:
 \begin{align}
u_i^{l}[t+1] &= \lambda (1-o_i^{l}[t])u_i^{l}[t] + \sum_{j=1}^{M_l}w_{ji}^{l}o_j^{l-1}[t]\\
o_i^{l}[t+1] &= g(u_i^{l}[t+1])
\label{lif2}
\end{align}
$\lambda = 1 - \frac{1}{\tau}$, and the function g is the threshold function. $w_{ji}^l$ is the synaptic weight from the $l^{th}$ layer from neuron $j$ to neuron $i$. $o_j^{l}[t]$ denotes the neuron $j$ spikes in $l^{th}$ layer at time $t$.
\begin{equation}
	g(u)=\left\{
		\begin{aligned}
			1, u > u_{th} \\
			0, u \leq u_{th}
		\end{aligned}
	\right.
	\label{lif3}
\end{equation}

\subsection{Spatio-temporal characteristics of SNNs}
The discontinuity of the spiking mechanism in the spiking neurons makes it challenging to apply the chain rule that connects the backpropagation gradient between the neural output and the neural input. In recent years, surrogate gradient~\citep{wu2018spatio,wu2019direct,neftci2019surrogate, bellec2020solution,fang2021deep} has been proposed to replace the discontinuous gradient with a smooth gradient function to enable the SNNs to conduct backpropagation in the spatial and temporal domains. Here we use the mean average firing rates of the last layer to approximate the classification label and train the network through the mean squared error (MSE)
:
\begin{equation}
	l = \frac{1}{S}\sum_{s=1}^S ||y_s - \frac{1}{T}\sum_{t=1}^To_t||^2
	\label{lif4}
\end{equation}
By applying chain rule, we can obtain the gradient with respect to weight:
 \begin{align}
\frac{\partial L}{\partial w^l} &= \sum_{t=1}^T \frac{\partial L}{\partial o_{i}^{l}[t]}\frac{\partial o_{i}^{l}[t]}{\partial u_{i}^{l}[t]}\frac{\partial u_{i}^{l}[t]}{w^l} \notag \\&= \sum_{t=1}^T \delta_i^l[t]g^{'}(u_{i}^{l}[t]) o^{l-1}[t]
\label{lif5}
\end{align}
$\delta_i^l[t] = \frac{\partial L}{\partial o_{i}^{l}[t]}$ denote the derivative with respect to $o$ in the $l^{th}$ layer at time step t, and can be derived from the $(l+1)^{th}$ layer (spatial) and $t+1$ time step (temporal). 
%\begin{align}
%\delta_i^l[t] &=\sum_{q=t}^{t_i}\sum_{j=1}^M \frac{\partial L}{\partial o_{j}^{l+1}[q]}\frac{\partial o_{j}^{l+1}[q]}{\partial o_{i}^{l}[t]} \notag \\&+ \frac{\partial L}{\partial o_{i}^{l}[t+1]}\frac{\partial o_{i}^{l}[t+1]}{\partial o_{i}^{l}[t]}
%\label{lif6}
%\end{align}
\begin{align}
\delta_i^l[t] =\sum_{j=1}^M \frac{\partial L}{\partial o_{j}^{l+1}[t]}\frac{\partial o_{j}^{l+1}[t]}{\partial o_{i}^{l}[t]} + \frac{\partial L}{\partial o_{i}^{l}[t+1]}\frac{\partial o_{i}^{l}[t+1]}{\partial o_{i}^{l}[t]}
\label{lif6}
\end{align}

\begin{figure}[htbp]
\centering
\includegraphics[width=8.5cm]{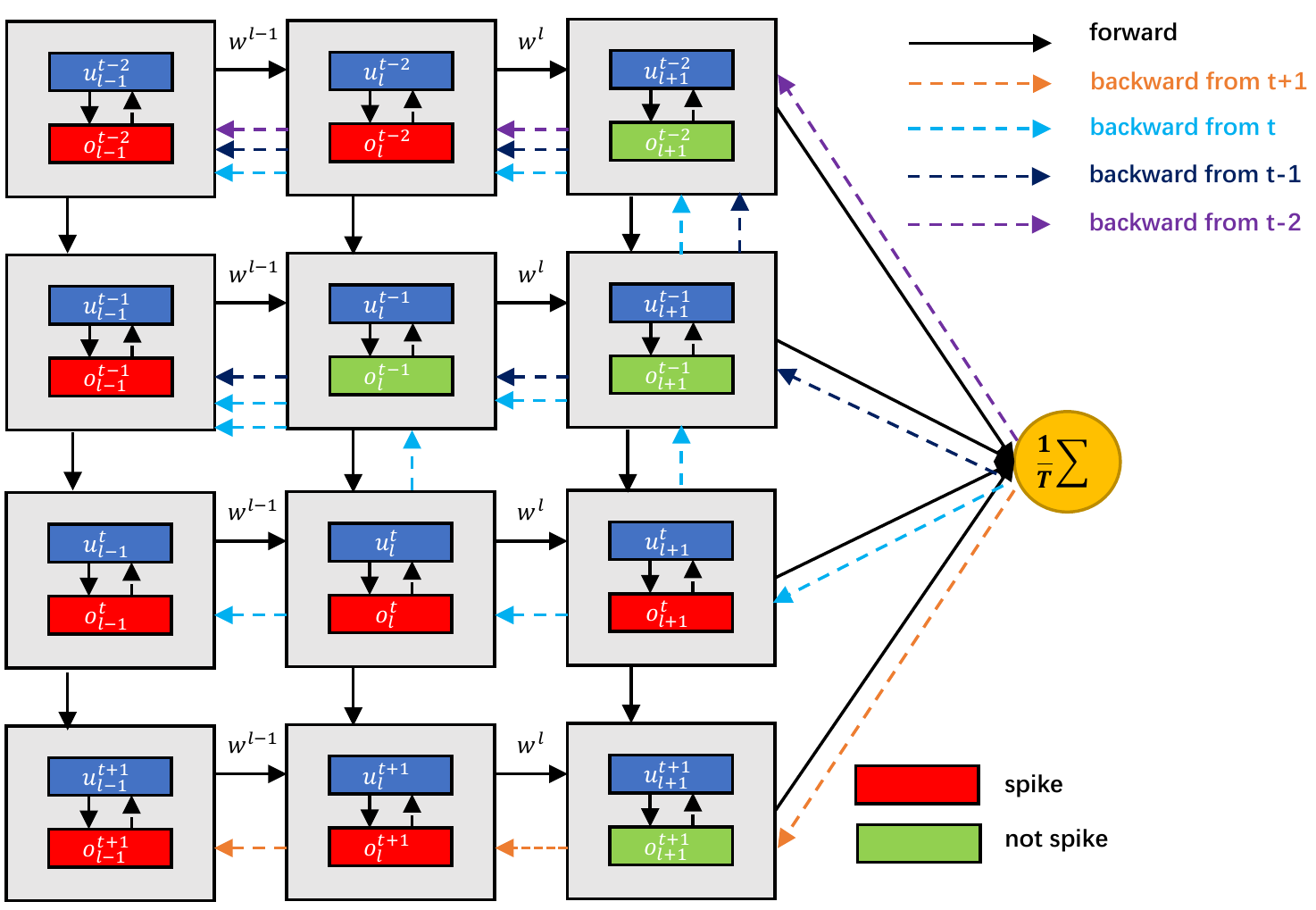}
\caption{The forward and backward process of spiking neural networks. The dotted lines of different colors indicate the impact on the network at different time steps, and the earlier the spike time, the greater the impact on the network.}
\label{fig1}
\end{figure}

As can be seen in Eq.~\ref{lif6}, due to the chain rule of the temporal dimensions, the gradient of the current moment will be affected by all the subsequent moments, and the details can be seen in Fig.~\ref{fig1}. 
Different colors indicate the backpropagation effect at different time steps. 
The earlier spiking $o_t$ repeatedly participates in the weight update, indicating that the farther away from the spiking time is, the greater the impact on the weight. 
While in neurophysiology, the farther away the spiking activity is from the current moment, the smaller the effect. Frequent participation in the calculation of the gradient will lead to the increase of synaptic strength so that neurons will frequently emit spikes, which will increase the overall energy consumption of the network. 
Therefore, we propose a biological plausible spatial adjustment (BPSA) to control better the update of the network weight by the error at each moment.

%这边可以加个STDP的图

The membrane potential of the spiking neuron depends on the membrane potential at the previous moment, which also allows the error to propagate back along the temporal dimension.
\begin{align}
\frac{\partial o_{i}^{l}[t+1]}{\partial o_{i}^{l}[t]} &= \frac{\partial o_{i}^{l}[t+1]}{\partial u_{i}^{l}[t+1]}\frac{\partial u_{i}^{l}[t+1]}{\partial u_{i}^{l}[t]}\frac{\partial u_{i}^{l}[t]}{\partial o_{i}^{l}[t]}  \notag \\
&= g^{'}(u_i^l[t+1])\lambda (1-o_i^l[t])\frac{1}{g^{'}(u_i^l[t])}
\label{equ_lif_6}
\end{align}

\begin{figure}[htbp]
\centering
\includegraphics[width=9cm]{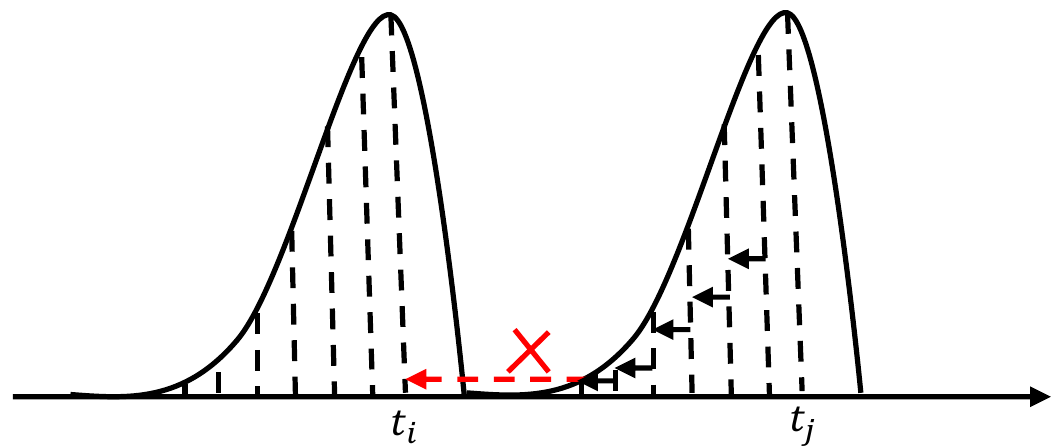}
\caption{The temporal backpropagation of LIF neurons can only within single spike period, and cannot propagate cross spikes.}
\label{fig2}
\end{figure}

As can be seen in Eq.~\ref{equ_lif_6}, when last moment spikes, $o_i^l[t]=1$, the membrane potential will be reset to 0. The temporal error is clipped and cannot continue to propagate. The temporal error of a spiking neuron only propagates within a single spike period and cannot propagate forward across multiple spikes, which can also be seen in Fig.~\ref{fig2}. The derivation of the current spikes along the temporal dimension can only be transmitted to the moment when the spike occurred last time and cannot continue to propagate forward.

\section{Methods}
To tackle the problems mentioned above, in this section, we propose the Biological Plausible Spatial and Temporal adjustment to regulate better the error propagates along the spatial and temporal dimensions.
\subsection{Biological Plausible Spatial Adjustment}
The membrane potential of spiking neurons changes as a process of information accumulation. After the neurons have accumulated enough information, they will send the information to the post-synaptic neurons in the form of spikes. As a result, the binary spikes can be regarded as a normalization of the information contained in the membrane potential. Therefore, for the backpropagation, it is more reasonable to only calculate the gradient of the neuron at the moment of spiking to the membrane potential.  As a result, this paper proposes a method for the biological plausible spatial adjustment. When the membrane potential does not reach the threshold, we will clip the gradient of the spikes to the membrane potential to avoid the problems of repeated updates at an earlier time in Fig.~\ref{fig1}. When the membrane potential reaches the threshold, we normalize the membrane potential and spread the information in the form of spikes. Then the derivative of the spikes concerning the membrane potential can be expressed as Eq.~\ref{equ_lif_7}.

\begin{equation} 
	\frac{\partial o_{i}^{l}[t]}{\partial u_{i}^{l}[t]} = \left\{
		\begin{aligned}
			&\frac{1}{u_{i}^{l}[t]}, \quad s^{i}_{t, n} = 1 \\
			&0, \quad otherwise
		\end{aligned}
	\right.
	\label{equ_lif_7} 
\end{equation}
This method can consider the influence of the spikes generated by the membrane potential of different strengths on the parameter update in the backpropagation process. For a spike excited by larger membrane potential, there will be a minor optimization step for the model parameters in the backpropagation process to ensure the stability of the spikes. The spikes excited by the membrane potential near the spike threshold $u_{th}$ will have a more significant impact on the model parameters, allowing the model to quickly push the membrane potential close to the threshold away to obtain more stable spikes. 
 
\subsection{Biological Plausible Temporal Adjustment}
In biological neurons, the spike the neuron fires will affect the subsequent spikes of the neuron. When the traditional backpropagation through time optimizes the parameters of the SNNs, the gradient of the loss function to the neuron output will only be propagated from the time the neuron was last excited to the present and will not cross the spikes as shown in Eq.~\ref{equ_lif_6} and Fig.~\ref{fig1}. This leads to the fact that the neurons in the SNN trained by the BPTT algorithm cannot consider the influence between spikes in the temporal domain. And this paper proposes a biological temporal adjustment cross the spikes. 

\begin{figure}[htbp]
\centering
\includegraphics[width=4cm]{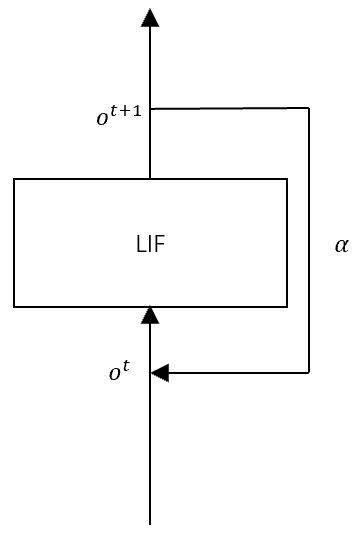}
\caption{The temporal residual pathway helps the error transfer from time step $t+1$ to time step $t$.}
\label{fig3}
\end{figure}

As can be seen in Fig.~\ref{fig3}, we add a temporal residual feedback pathway to propagate the temporal errors across spikes, and the error can be written as:
\begin{align}
\frac{\partial o_{i}^{l}[t+1]}{\partial o_{i}^{l}[t]} &= g^{'}(u_i^l[t+1])\lambda (1-o_i^l[t])\frac{1}{g^{'}(u_i^l[t])} \notag\\
&+ \lambda \alpha g^{'}(u_i^l[t+1])
\label{equ_lif_8}
\end{align}

As can be seen in Eq.~\ref{equ_lif_8}, we introduce a residual factor $\alpha$ to control t he error transfer from time step $t+1$ to $t$. 

%我们在时序的反向通路上添加了
\begin{figure}[htbp]
\centering
\includegraphics[width=8.5cm]{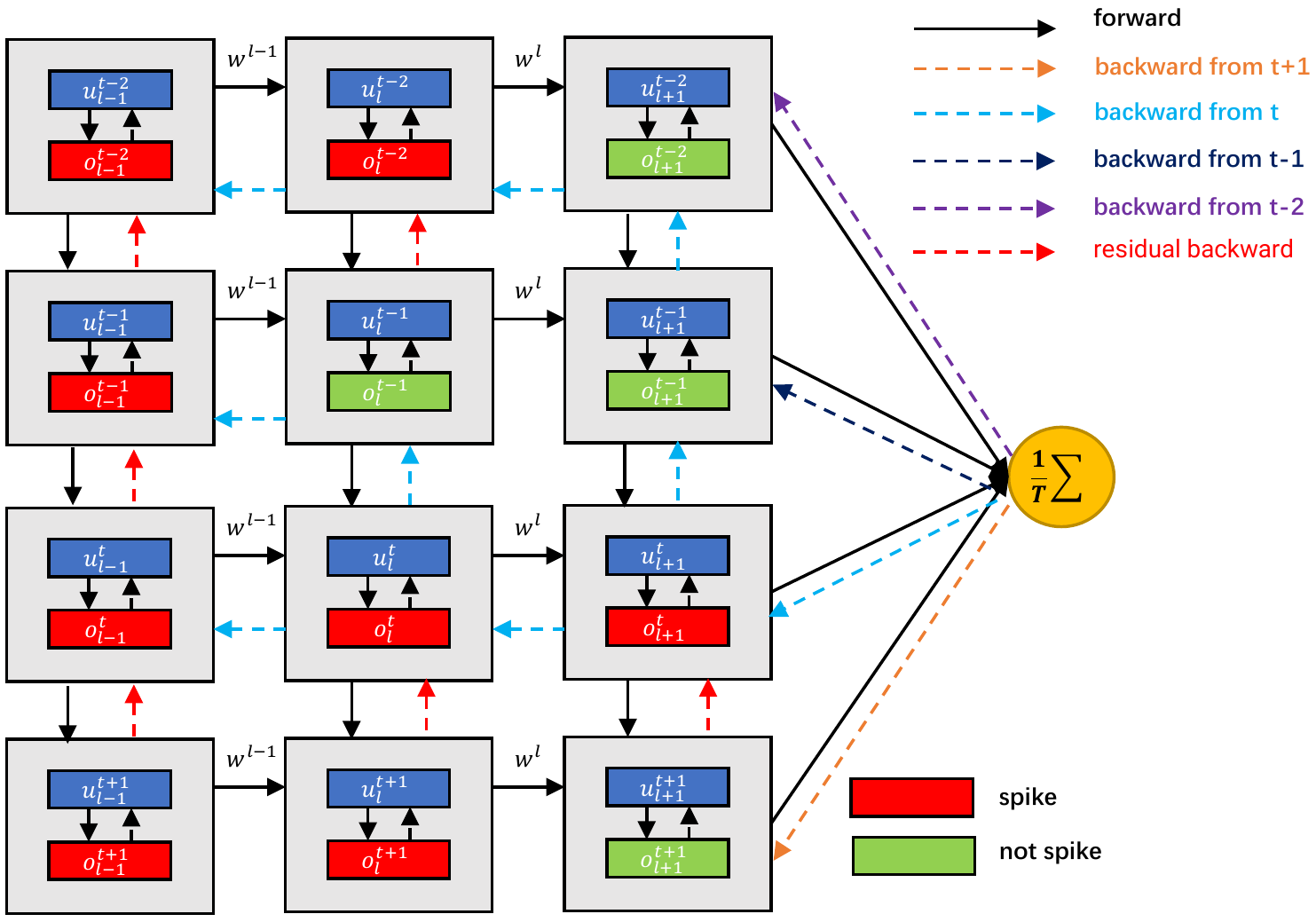}
\caption{The while forward and backward of our method.}
\label{fig11}
\end{figure} 

As can be seen in Fig.~\ref{fig11}, with the introduction of the biological spatial and temporal adjustment, the influence of different spikes become more reasonable, and the temporal residual backward pathway enables it to propagate errors over spikes.  In the subsequent experiments, $\alpha$ is set with 0.8. 
\section{Experiment}
In this section, we conducted experiments using the PyTorch framework\citep{paszke2017automatic} 
on NVIDIA A100 GPU with several datasets to verify our method. The weight of the network uses the default initialization method of PyTorch. We use the AdamW\citep{loshchilov2017decoupled} as the optimizer, the learning rate $lr$ is set with $1\times 10^{-3}$, and the same learning rate control strategy as in SGDR~\citep{loshchilov2016sgdr} is used. At the beginning of training, the same method in TSSL-BP is used to warmup the model. The membrane potential threshold $u_{th}$ of the neuron is set to 0.5, the membrane potential decay constant $\lambda=0.9$, and the default simulation duration T is set to 16.  The training epochs are set at 300. First, we conduct experiments on the static MNIST~\cite{lecun1998mnist} and CIFAR10~\cite{Krizhevsky09learningmultiple} datasets and then conduct experiments on the neuromorphic datasets N-MNIST~\cite{orchard2015converting}, DVS-Gesture~\citep{amir2017low} and DVS-CIFAR10~\citep{li2017cifar10}. For the static datasets, we use the direct input encoding used in the ~\citep{wu2019direct}, as well as the voting strategy. For the neuromorphic dataset N-MNIST and DVS-CIFAR10, we use the same data preprocessing  strategy used in SpikingJelly~\citep{fang2020other}. For different datasets, we designed three different network structures to adapt different sizes and complexity As shown in Tab.~\ref{structure}, DP denotes neuron dropout~\citep{lee2020enabling}, and C denotes the Conv-BN-ReLU-LIF operation.

\begin{table}[h]

	\centering
	\caption{Different network structures}
	\begin{tabular}{cl}
		\toprule[2pt]
		Small  & 128C3-MP2-128C3-256C3-MP2\\&-4096FC-DP-10Voting \\
		\hline 
		Middle & 128C3-MP2-128C3-MP2-256C3-MP2\\
		&-512C3-AP4-512FC-10Voting \\
		\hline
		Large  & 128C3-128C3-AP2-128C3-AP2-256C3-AP2\\&-512C3
			   -AP2-1024C3-AP4-DP-1024FC-10Voting \\
		\bottomrule[2pt]
	\end{tabular}
	\label{structure}
\end{table}
%\begin{table}[h]
%\small
%	\centering
%	\caption{Performances of BP based SNNs on MNIST}
%	\begin{tabular}{c|c|c|c|c}
%		\toprule[2pt]
%		Models & Time & Mean(\%) &Std(\%) &Best  (\%) \\
%		\hline%\midrule%[2pt]
%		Spike CNN~\shortcite{lee2016training} & $>200$& - & - & 99.31\\
%		SLAYER~\shortcite{shrestha2018slayer} & 300 & 99.36 & 0.05 & 99.41\\
%		STBP~\citep{wu2018spatio} & $>100$  & - & - & 99.42\\
%		
%		HM2BP~\shortcite{jin2018hybrid} & 400 & 99.42 & 0.11 & 99.49\\
%		LISNN~\citep{cheng2020lisnn} & 20 & - & - & 99.5\\
%		TSSL-BP~\citep{zhang2020temporal} & 5 & 99.50 & 0.02 & 99.53\\
%		ST-RSBP~\citep{zhang2019spike} & 400 & 99.57 & 0.04 & 99.62\\
%		Our Method 															  & 16&  \textbf{99.58} &  0.06 &  \textbf{99.67}  \\
%		%BackEISNN\shortcite{zhao2021backeisnn} 															  & 20&  99.58 &  0.06 &  99.67  \\
%		
%		\bottomrule[2pt]
%	\end{tabular}
%	\label{static}
%\end{table}
\begin{table}[h]
\small
	\centering
	\caption{Performances of BP based SNNs on MNIST}
	\begin{tabular}{cccc}
		\toprule[2pt]
		Models  & Mean(\%) &Std(\%) &Best  (\%) \\
		\hline%\midrule%[2pt]
		Spike CNN~\shortcite{lee2016training} & - & - & 99.31\\
		SLAYER~\shortcite{shrestha2018slayer} & 99.36 & 0.05 & 99.41\\
		STBP~\citep{wu2018spatio}    & - & - & 99.42\\
		
		HM2BP~\shortcite{jin2018hybrid}  & 99.42 & 0.11 & 99.49\\
		LISNN~\citep{cheng2020lisnn}  & - & - & 99.5\\
		TSSL-BP~\citep{zhang2020temporal}  & 99.50 & 0.02 & 99.53\\
		ST-RSBP~\citep{zhang2019spike}  & 99.57 & 0.04 & 99.62\\
		Our Method 															  &  \textbf{99.58} &  0.06 &  \textbf{99.67}  \\
		%BackEISNN\shortcite{zhao2021backeisnn} 															  & 20&  99.58 &  0.06 &  99.67  \\
		
		\bottomrule[2pt]
	\end{tabular}
	\label{static}
\end{table}
\subsection{MNIST}
MNIST is one of the most common classification datasets in the deep learning domain, with 60,000 training datasets and 10,000 test datasets. The samples in the datasets are 28*28 gray-scale images representing handwritten numbers from 0 to 9, respectively. We use the small structure for the evaluation. We compare our model with several state-of-the-art snn models. The experiment was repeated five times and can be seen from the Tab.~\ref{static}, our method obtains a relatively high performance compared to other BP-based SNN models.
\subsection{CIFAR10}
The CIFAR10 dataset is a more challenging dataset for most existing SNNs. The training set has 50,000 samples, and the test set has 10,000 samples. The dataset is a 32*32 color dataset. Deeper network to achieve better network effect. The network structure is the middle structure. We compare our result with several deep SNN models, including the conversion methods and bp-based methods. STBP NeuNorm is the STBP method with the Neuron Norm.  
\begin{table}[h]
\small
	\centering
	\caption{Performances of BP and conversion absed SNNs on CIFAR10}
	\begin{tabular}{ccc}
		\toprule[2pt]
		Models & Method & ACC(\%)  \\
		\hline%\midrule%[2pt]
		Spiking CNN~\shortcite{hunsberger2015spiking}			&  Conversion    & 82.95 \\
		
		ContinueSNN~\citep{rueckauer2017conversion}  &  Conversion     & 88.82 \\
		Spike-Norm~\citep{sengupta2019going}  				&  Conversion   & 91.55 \\
		STBP~\citep{wu2019direct} & BP & 89.83\\
		STBP NeuNorm~\citep{wu2019direct} & BP & 90.53\\
		BackEISNN~\citep{zhao2021backeisnn} & BP & 90.93\\
		SBPSNN~\citep{lee2020enabling}			& BP & 90.95 \\
		TSSL-BP~\citep{zhang2020temporal} & BP & 91.41\\
		
		Our Method 															& BP& \textbf{92.15}  \\

		\bottomrule[2pt]
	\end{tabular}
	\label{static2}

\end{table}

As can be seen in Tab.~\ref{static2}, our method reaches the state-of-the-art performance, demonstrating the superiority of our models. 

\subsection{N-MNIST}
N-MNIST is the neuromorphic version of MNIST. The Dynamic Version Sensor (DVS) is put in front of the static images on a computer screen. The images shift due to the DVS moved in the direction in three sides of the isosceles triangle in turn, and the two-channel spike event (on and off) is collected. The network structure is set the same as that in the CIFAR10 experiment.  %这里需要添加如何对于N-MNIST处理的

\begin{table}[h]
\small
	\centering
	\begin{threeparttable}
	\caption{Performances of BP based SNNs on N-MNIST}
	\begin{tabular}{ccc}
		\toprule[2pt]
		Models & Method & ACC(\%)  \\
		\hline%\midrule%[2pt]
		HM2-BP~\cite{jin2018hybrid} &400-400 & 98.88 \\
		SLAYER~\cite{shrestha2018slayer} &Net 1 & 99.2 \\
		TSSL-BP 30~\cite{zhang2020temporal} &Net 1 & 99.28 \\
		IIRSNN~\cite{fang2020exploiting} &Net 1 & 99.28 \\
		TSSL-BP 100~\cite{zhang2020temporal} &Net 1& 99.4 \\
		STBP~\cite{wu2019direct} & Net 2 & 99.44 \\
		LISNN~\cite{cheng2020lisnn} &Net 3 &  99.45 \\
		STBP NeuNorm~\cite{wu2019direct} &Net 2& 99.53 \\
		BackEISNN~\cite{zhao2021backeisnn} &Net 1& 99.57 \\
		Our Method 															& Middle& \textbf{99.71}\\		
		\bottomrule[2pt]
	\end{tabular}
	\label{static3}
	\begin{tablenotes}    %这行要添加， 从这开始
        \footnotesize               %这行要添加
        \item[1] Net 1 is 12C5-P2-64C5-P2
        \item[2] Net 2 is 128C3-128C3-P2-128C3-256C3-P2-1024-Voting
        \item[3] Net 3 is 32C3-P2-32C3-P2-128 
      \end{tablenotes}  
\end{threeparttable} 
\end{table}
As can be seen in Tab.~\ref{static3}, our method has surpassed STBP by 0.3\%, even with the introduction of NeuNorm, our work still performs better than them, which fully demonstrates the superiority of our work. 
\subsection{DVS-Gesture}
DVS-Gesture is a real-time gesture recognition dataset reported by DVS. The dataset has 11 hand gestures such as hand clips, arm roll, etc., collected from 29 individuals under three illumination conditions. We compare our models with several currently ANNs and SNNs due to their wealthy temporal information. The network was set with the large structure. 

\begin{table}[h]
	\centering
	\caption{Performances of several SNNs and DNNs on DVS-Gesture}
	\begin{tabular}{ccc}
		\toprule[2pt]
		Models & Method & ACC(\%)  \\
		\hline%\midrule%[2pt]
		SLAYER~\shortcite{shrestha2018slayer} & SNN & 93.64\\
		STBP-tdBN~\cite{zheng2021going} &SNN & 96.87 \\
		PointeNet~\cite{wang2019space} &DNN & 97.08\\
		RG-CNN~\cite{bi2020graph} &DNN & 97.2\\
		LMCSNN~\cite{fang2020incorporating} & SNN & 97.57 \\ 
		STFilter~\cite{ghosh2019spatiotemporal} &DNN & 97.75\\
		Our Method 															& SNN& \textbf{98.96}  \\

		\bottomrule[2pt]
	\end{tabular}
	\label{static4}
\end{table}

As can be seen in Tab.~\ref{static4}, for the more complex gesture dataset, our model surpasses the latest STBP-tdBN by 2\%. Even compared with the traditional DNNs, our model far exceeds them and has reached state-of-the-art performance compared with other current famous SNNs and DNNs. 
 \begin{table*}[h]

	\centering
	\caption{The energy efficiency study of our model with baseline on different datasets }
	\begin{tabular}{cccc}
		\toprule[2pt]
		 Dataset 		&  Accuracy   	&  Firing-rate  &EE = $\frac{E_{ANN}}{E_{SNN}}$ \\
		\midrule[2pt]
		 MNIST   		&  99.58\%/99.42\%	& 0.082/0.183	& 35.1x/15.7x	 
		\\
		
		  N-MNIST 		&  99.61\%/99.32\%  & 0.097/0.176   & 29.6x/16.3x		    \\
		
		 CIFAR10 		&  92.33\%/89.49\%  & 0.108/0.214	& 26.6x/13.4x				\\ 
		
		 DVS-Gesture 	&  98.26\%/93.92\%  & 0.083/0.165   & 34.6x/17.4x				\\ 
							
								DVS-CIFAR10  &  77.76\%/71.40\%  & 0.097/0.177  & 29.5x/16.2x				\\
		\bottomrule[2pt]
	\end{tabular}
	\label{compare2}
\end{table*} 
\begin{figure*}[h]
	\centering
	\begin{minipage}{5cm}
		\includegraphics[width=5cm]{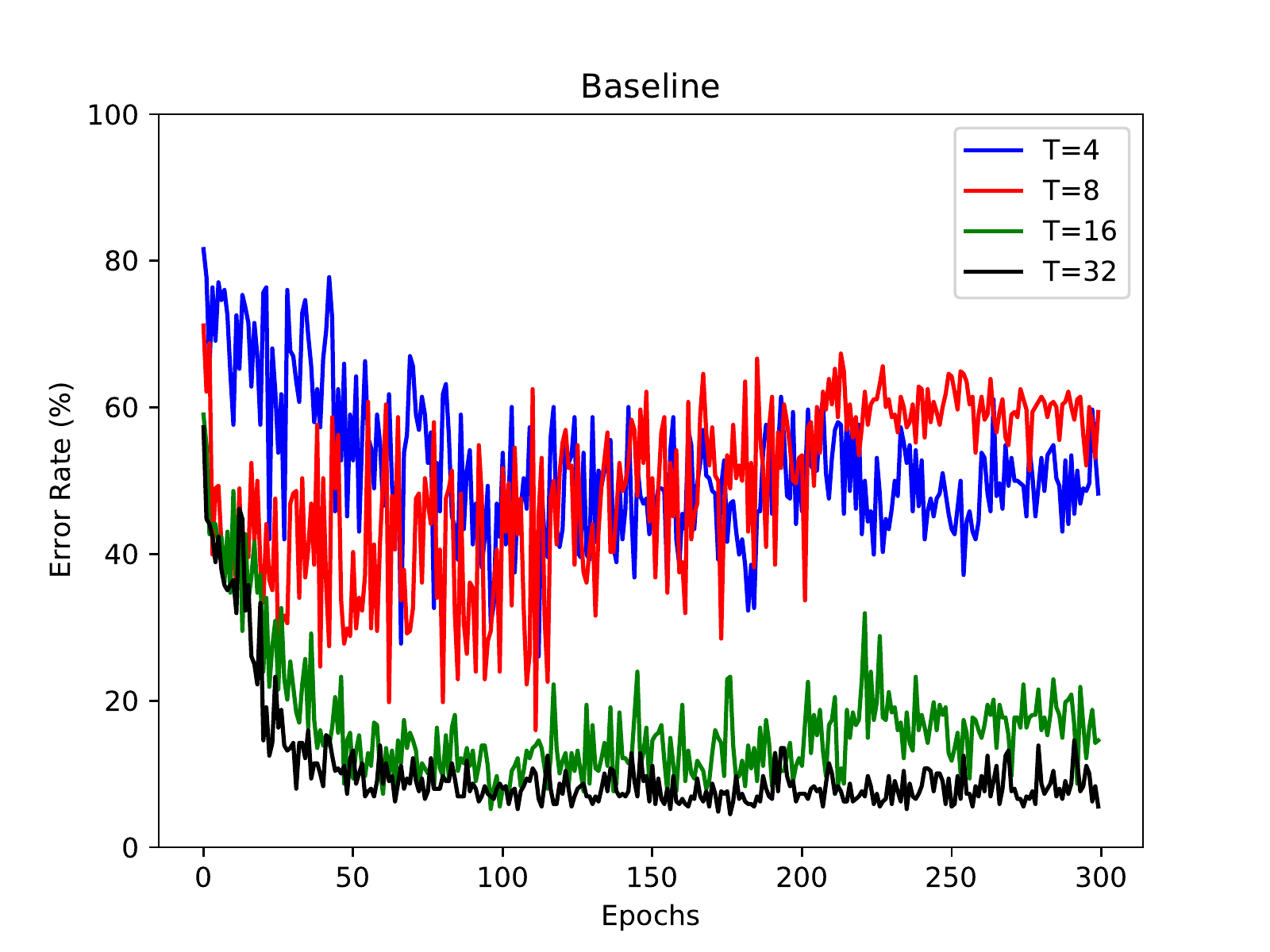}
		%\label{1}
		%\caption{a}
	\end{minipage}
	\begin{minipage}{5cm}
		\includegraphics[width=5cm]{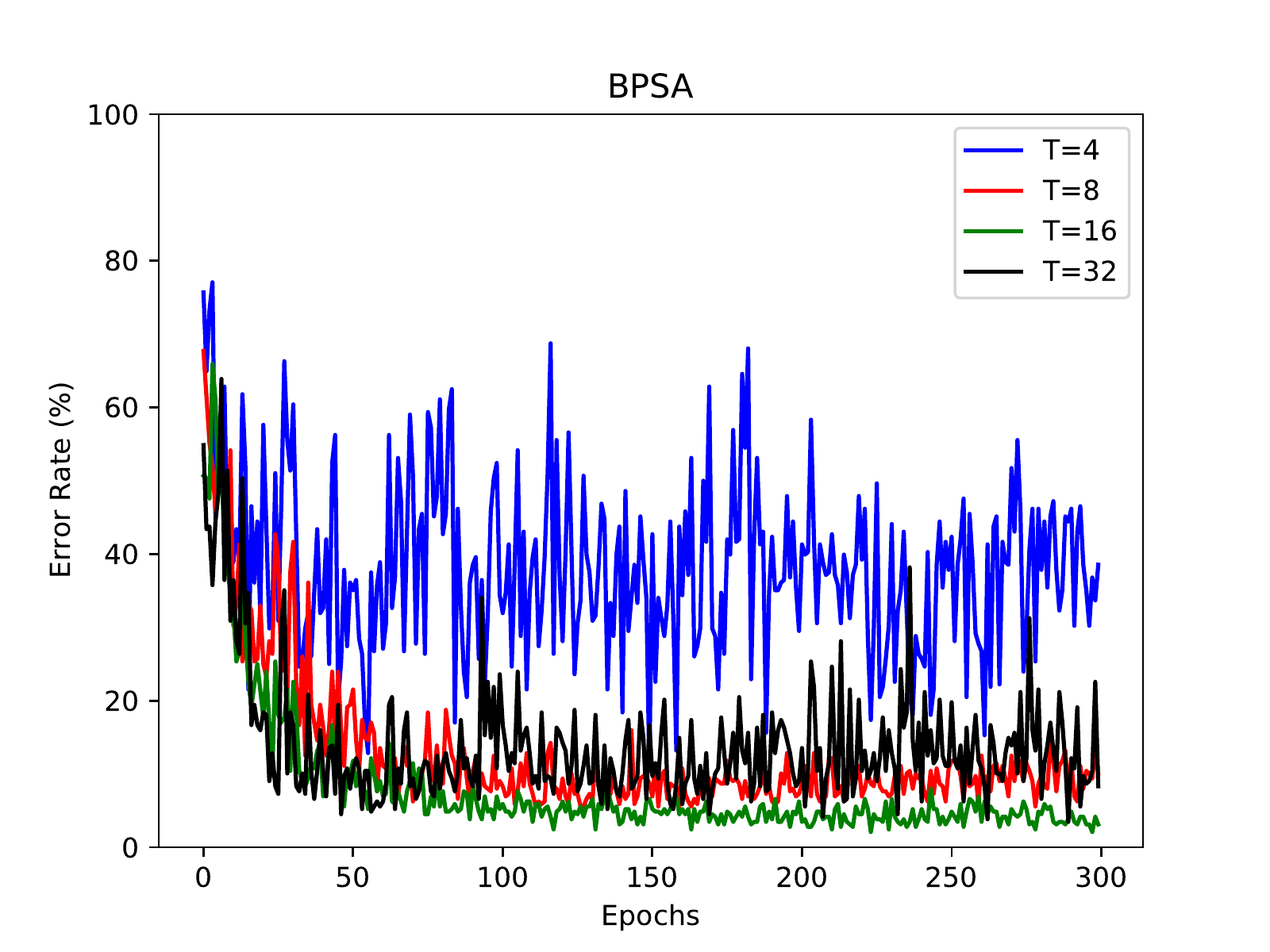}
		%\caption{b}
	\end{minipage}
	\begin{minipage}{5cm}
		\includegraphics[width=5cm]{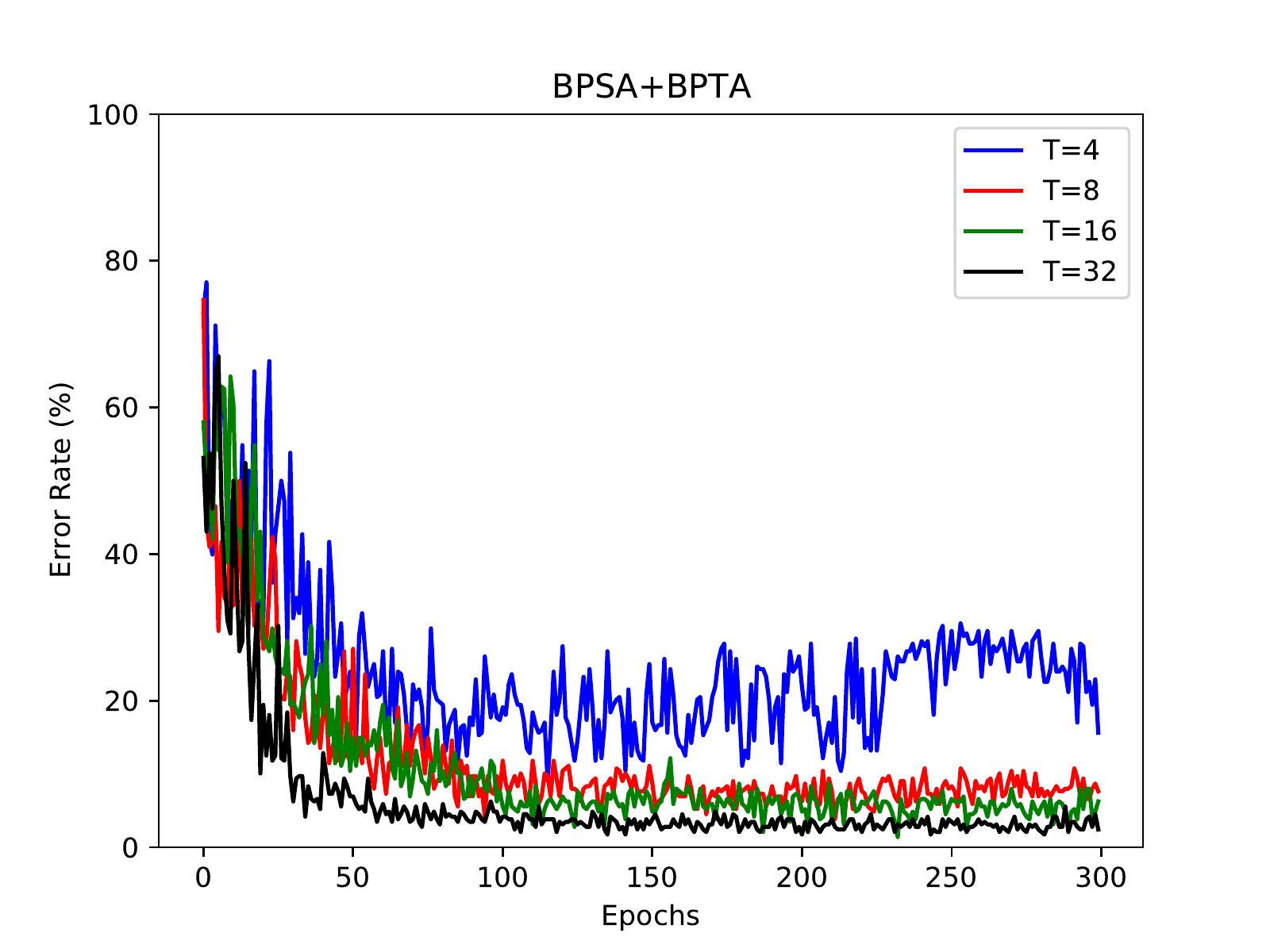}
		%\caption{b}
	\end{minipage}
	\caption{The test accuracy of different simulation lengths on DVS-Gesture dataset with our method and the baseline.}
	\label{ACC}
\end{figure*} 
 
\subsection{DVS-CIFAR10}
DVS-CIFAR10 is a neuromorphic version converted from the CIFAR10 dataset. 10,000 frame-based images are converted into 10,000 event streams with DVS, an important benchmark for comparison in SNN domains. The network was set with the large structures. 
\begin{table}[h]
\small
	\centering
	\caption{Performances of BP based SNNs on MNIST}
	\begin{tabular}{ccc}
		\toprule[2pt]
		Models & Method & ACC(\%)  \\
		\hline%\midrule%[2pt]
		STBP NeuNorm~\citep{wu2019direct} &SNN & 60.5 \\ 
		STBP-tdBN~\citep{zheng2021going} &SNN & 67.8 \\ 
		ASF-BP~\citep{wu2021training} &SNN & 58.2\\
		TandomSNN~\citep{wu2021tandem} &SNN & 65.59\\ 
		RG-CNN~\citep{bi2020graph} &DNN & 54\\
		LMCSNN~\citep{fang2020incorporating} & SNN & 74.8 \\ 
		Dart~\citep{ramesh2019dart} & DNN &65.78\\
		Our Method 															& SNN& \textbf{78.95}  \\

		\bottomrule[2pt]
	\end{tabular}
	\label{static5}
\end{table}

As can be seen in Tab.~\ref{static5}, DVS-CIFAR10 is a more complex neuromorphic dataset, which contains wealthy spatio-temporal information. Compared with the latest STBP-tdBN, we surpassed them by nearly 11\%. For LMCSNN~\citep{fang2020incorporating}, which make many parameters in the LIF spiking neurons learnable, we also surpass them by 4\%. Our method has achieved state-of-the-art performance for the DVS-CIFAR10 dataset.  

\section{Discussion}

In this section, firstly, we conduct the ablation study to the BPSA and BPTA mentioned above and analyze the contribution of each module. Secondly, we will explore the energy consumption of the spiking neural networks for these adjustments. Finally, we discuss the latency of the SNNs affected by these adjustments. Through the analysis, it is fully illustrated that the above two modules can make the behavior of the spiking neurons more stable and establish a better performance while reducing the network latency and energy consumption.
\subsection{Ablation Study}
We conduct an ablation study on the neuromorphic datasets DVS-Gesture and DVS-CIFAR10 due to the more complex spatial structure and stronger temporal information, fully illustrating our adjustments' importance. We use~\citep{wu2018spatio} as our baseline and then continue to add the biological spatial and temporal adjustments. 

%剥离分析
%能效分析
%延迟分析
\begin{table}[h]
	\centering
	\caption{The ablation study of the two adjustments on DVS-Gesture and DVS-CIFAR10 dataset}
	\begin{tabular}{c c c c}
		\toprule[2pt]
		 				 	&  Baseline 		&  BPSA   		&  BPSA+BPTA   \\
		\hline
		DVS-Gesture  		&  93.92   	&  97.56	& 98.96	 \\
		DVS-CIFAR10  		&  71.40	&  75.30	& 78.95  \\
		\bottomrule[2pt]
	\end{tabular}
	\label{ablation1}
\end{table}

As can be seen in Tab.~\ref{ablation1}, with the introduction of the two adjustments, the performance of the network is gradually improving, among which the spatial adjustments bring more significant improvement. 

\begin{figure}[h]
	\centering
	\includegraphics[width=0.5\textwidth]{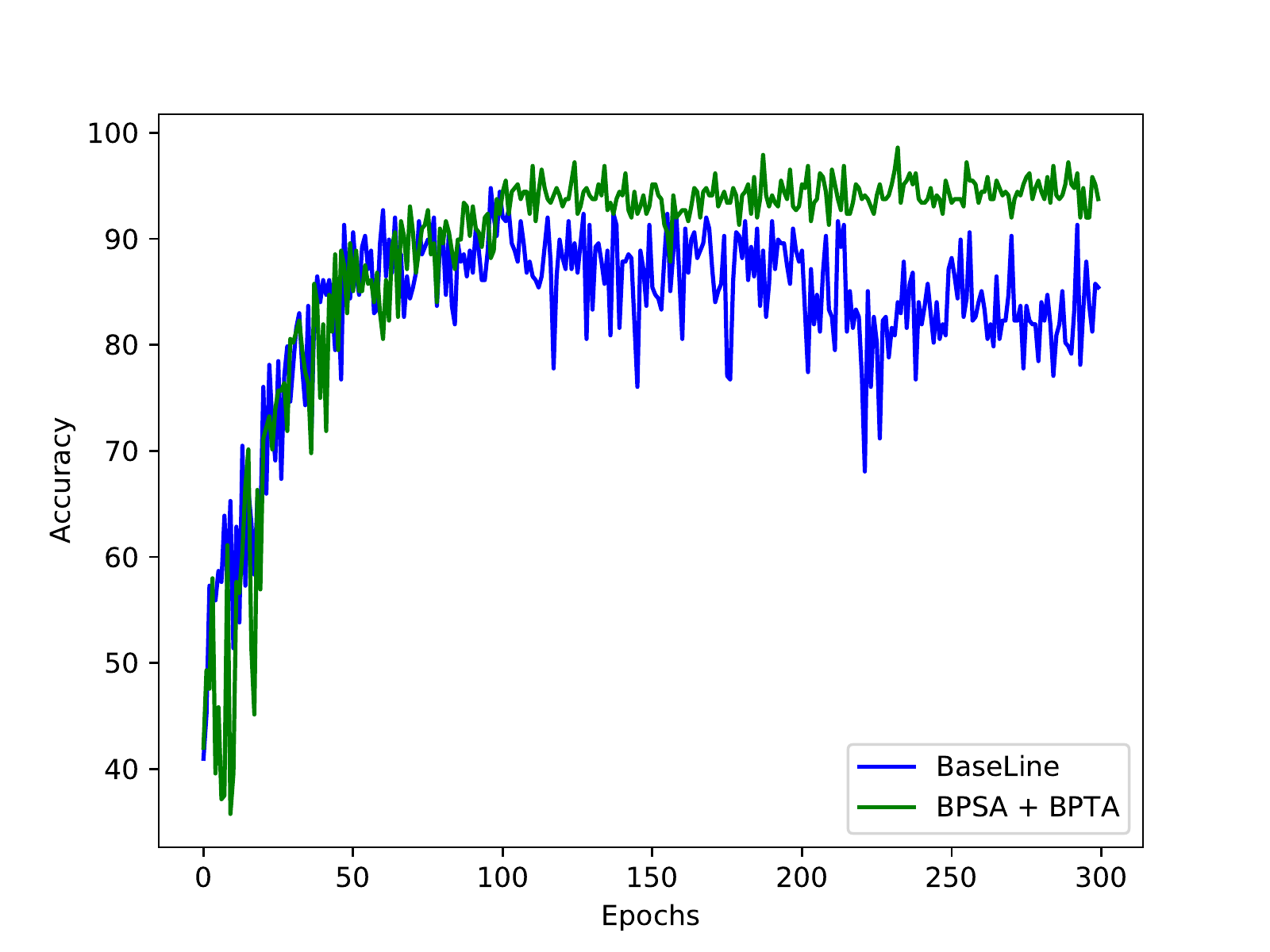}
	\caption{The test accuracy curve on DVS-Gesture of our method and the baseline}
	\label{training}
\end{figure}

We also give the test curves of the DVS-Gesture dataset. As shown in Fig.~\ref{training}, with the number of training epochs increases, the accuracy of the model with biological spatial-temporal adjustments fluctuates less. Because with the introduction of the two adjustments, the firing pattern of neurons is more stable, making the model more robust to more minor parameter changes. Meanwhile, a reasonable gradient allocation strategy in the gradient backpropagation improves the model's generalization performance and avoids overfitting to a certain extent.

\subsection{Energy Efficiency Study} 
We compare the accuracy and the energy efficiency of the SNNs trained by the method used in~\citep{wu2018spatio}, the model we propose and the ANNs using the same network structure and network parameters. Most operations in ANNs are multiply-accumulate (MAC), while in SNNs the spikes transmit in the network are sparse, and the spikes are integrated into the membrane potential. As a result, most operations in SNNs are accumulate (AC) operations. 
We calculate the energy consumption of the spiking neural network by multiplying FLOPS (floating-point operations) and the energy consumption of MAC and AC operations. We use the same energy efficiency method used in~\citep{chakraborty2021fully}. And the computation details can be seen in Eq.~\ref{equ_lif_11}.

\begin{align}
E_{ANN} &= FLOPS_{ANN} \times E_{FL\_MAC}  \notag \\
E_{SNN} &= FLOPS_{SNN} \times E_{INT\_AC} \times T
\label{equ_lif_11}
\end{align}

As can be seen in Tab~\ref{compare2}, our method has a lower firing rate, and higher energy efficiency. The training method of the SNNs proposed in this paper distributes the gradient more reasonably along the spatial and temporal dimension, avoiding the problem that the earlier spiking neurons would have a more significant influence on the network parameters. As well, the cross spikes propagation will enhance the temporal dependence of the SNNs. Therefore, the method proposed in this paper achieves lower network power consumption while maintaining a higher accuracy. 

\subsection{Latency Study}
The latency of the SNNs is one of the main problems that restrict the development of SNNs. The spiking neurons need to accumulate membrane potential, and once they reach the threshold, they fire spikes and transmit information. Therefore, SNNs often require a long simulation time to achieve higher performance. Here we study the influence of different simulation lengths on the network performance.

\begin{table}[h]
	\centering
	\caption{The test accuracy on DVS-Gesture dataset of different simulation lengths of our method and the baseline}
	\begin{tabular}{ccccc}
		\toprule[2pt]
		 				 	&  T=32 	&  T=16   	&  T=8	&  T=4  \\
		\hline
		BPSA+BPTA  				&  98.27   	&  98.26	& 96.18	&  96.18 \\

		BPSA  					&  96.53	&  97.56	& 94.44 &  89.58 \\

		Baseline					&  95.49	&  93.92    & 84.03	&  73.96 \\
		\bottomrule[2pt]
	\end{tabular}
	\label{ttt}
\end{table}
As shown in the Fig.~\ref{ACC}, when our adjustment is not introduced, when the simulation time is reduced, the test curve of the network is not very smooth, that is, the network needs a long simulation time to converge. As can be seen in Tab.~\ref{ttt}, with the introduction of the two modules, our training method still achieves high accuracy while reducing the simulation time. The low latency of our approach further lays the foundation for the practical application of SNNs.
 
\section{Conclusion}
In this paper, first, we analyze the existing problems in the SNNs trained with BP. We find that the current setting will cause the earlier spiking neurons repeated participate in the gradient calculation of the network, making a more significant influence on the network weight. The BPTT algorithm on the SNNs only propagates errors backward in a single spike period. The temporal dependence between spikes will be truncated. By introducing the biological spatial adjustment, it will consider the spikes generated by the membrane potential of different strengths, which will have different effects on the parameter update during the backpropagation process. In addition, the temporal adjustment with temporal residual influence is introduced, which considers the backpropagation across the spikes. We have gotten remarkable performance on MNIST and CIFAR10 datasets and achieved the current best performance on N-MNIST, DVS-Gesture, and DVS-CIFAR10 datasets. By analyzing the energy consumption and latency of the SNNs, we find that the biological spatial and temporal adjustment significantly reduces the energy consumption and latency while improving the performance.
\bibliography{aaai22}

\end{document}